\documentclass[11pt]{article}
\usepackage{amsmath,epsfig,amssymb,amsthm}
\usepackage{graphicx,subfigure}
\usepackage{palatino}
\usepackage{microtype}
\usepackage{mathrsfs}
\usepackage{algorithm,algpseudocode}

\def\x{\boldsymbol{x}}
\def\y{\boldsymbol{y}}

\begin{document}

\title{Fast $O(1)$ bilateral filtering using \\ trigonometric range kernels}

\author{Kunal~Narayan~Chaudhury, Daniel~Sage, and Michael~Unser
\thanks{Correspondence: Kunal~N.~Chaudhury (kchaudhu@princeton.edu). Kunal N. Chaudhury is currently part of the Program in Applied and Computational Mathematics (PACM), Princeton University, Princeton, NJ 08544-1000, USA. Michael Unser and Daniel Sage are with the Biomedical Imaging Group, \'Ecole Polytechnique F\'ed\'erale de Lausanne, Station-17, CH-1015 Lausanne,  Switzerland. This work was supported by the Swiss National Science Foundation under grant 200020-109415.}}

\maketitle

\begin{abstract}
It is well-known that spatial averaging can be realized (in space or frequency domain) using algorithms whose complexity does not scale with the size or shape of the filter. These fast algorithms are generally referred to as constant-time or $O(1)$ algorithms in the image processing literature. Along with the spatial filter, the edge-preserving bilateral filter \cite{bilateralFilter} involves an additional range kernel. This is used to restrict the averaging to those neighborhood pixels whose intensity are similar or close to that of the pixel of interest. The range kernel operates by acting on the pixel intensities. This makes the averaging process non-linear and computationally intensive, especially when the spatial filter is large. In this paper, we show how the $O(1)$ averaging algorithms can be leveraged for realizing the bilateral filter in constant-time, by using trigonometric range kernels. This is done by generalizing the idea in \cite{bilateralFilter_fast} of using polynomial kernels. The class of trigonometric kernels turns out to be sufficiently rich, allowing for the approximation of the standard Gaussian bilateral filter. The attractive feature of our approach is that, for a fixed number of terms, the quality of approximation achieved using trigonometric kernels is much superior to that obtained in \cite{bilateralFilter_fast} using polynomials.
\end{abstract}

\section{Introduction}

\noindent The bilateral filtering of an image $f(\x)$ in the general setting is given by
\begin{equation*}
\tilde{f}(\x)=\eta^{-1} \int w(\x, \y) \ \phi(f(\x),f(\y)) \ f(\y) \ d\y
\end{equation*}
where
\begin{equation*}
\eta=\int w(\x, \y) \ \phi(f(\x),f(\y))  \ d\y.
\end{equation*}
In this formula, $w(\x,\y)$ measures the geometric proximity between the pixel of interest $\x$ and a nearby pixel $\y$. Its role is to localize the averaging to a neighborhood of $\x$. On the other hand, the function $\phi(u,v)$ measures the similarity between the intensity of the pixel of interest $f(\x)$ and its neighbor $f(\y)$. The normalizing factor $\eta$ is used to preserve constants, and in particular the local mean. 

In this paper, we consider the so-called \textit{unbiased form} of the bilateral filter \cite{bilateralFilter}, where $w(\x,\y)$ is translation-invariant, that is, $w(\x,\y)=w(\x-\y)$, and where the range filter is symmetric and depends on the difference of intensity, $\phi(f(\x),f(\y))=\phi(f(\x)- f(\y))$. In this case, the filter is given by
\begin{equation}
\label{BF}
\tilde{f}(\x)=\eta^{-1} \int_{\Omega} w(\y)  \phi(f(\x-\y)-f(\x))  f(\x-\y) \ d\y
\end{equation}
where
\begin{equation}
\label{normalization}
\eta=\int_{\Omega} w(\y) \phi(f(\x-\y)-f(\x))  \ d\y.
\end{equation}
We call $w(\x)$ the \textit{spatial kernel}, and $\phi(s)$ the \textit{range kernel}. The local support $\Omega$ of the spatial kernel specifies the neighborhood over which the averaging takes place. A popular form of the bilateral filter is one where both $w(\x)$ and $\phi(s)$ are  Gaussian \cite{bilateralFilter,bilateralFilter_fast,Baudes,Wang_bf}. 

The edge-preserving bilateral filter was originally introduced by Tomasi et al. in \cite{bilateralFilter} as a simple, non-iterative  alternative to  anisotropic diffusion \cite{PeronaMalik}. This was motivated by the observation that while standard spatial averaging performs well in regions with homogenous intensities, it tends to performs poorly in the vicinity of sharp transitions, such as edges. For the bilateral filter in \eqref{BF}, the difference $f(\x-\y)-f(\x)$ is close to zero in homogenous regions, and hence $\phi(f(\x-\y)-f(\x)) \approx 1$. In this case, \eqref{BF} simply results in the averaging of pixels in the neighborhood of the pixel of interest. On the other hand, if the pixel of interest $\x$ is in the vicinity of an edge, $\phi(f(\x-\y)-f(\x))$ is large when $\x-\y$ belongs to the same side of the edge as $\x$, and is small when $\x-\y$ is on the other side of the edge. As a result, the averaging is restricted to neighborhood pixels that are on the same side of the edge as the pixel of interest. This is the basic idea which allows one to perform smoothing while preserving edges at the same time. Since its inception, the bilateral filter has found widespread use in several image processing, computer graphics, and computer vision applications. This includes denoising \cite{Bennet}, video abstraction \cite{VideoAbstraction}, demosaicing \cite{Ramanath}, optical-flow estimation \cite{Xiao}, and stereo matching \cite{Yang_BLF_stereo}, to name a few. More recently, the bilateral filter has been extended by Baudes et al. \cite{Baudes} to realize the popular non-local neighborhood filter, where the similarity between pixels is measured using patches centered around the pixels.

The direct implementation of \eqref{BF} turns out to be rather computationally intensive for real time applications. Several efficient numerical schemes have been proposed in the past for implementing the filter in real time, even at video rates \cite{Durand,Weiss,Pham,Paris}. These algorithms (with the exception of \cite{Durand}), however, do not scale well with the size of the spatial kernel, and this limits their usage in high resolution applications. A significant advance was obtained when Porikli \cite{bilateralFilter_fast} proposed a constant-time implementation of the bilateral filter (for arbitrary spatial kernels) using polynomial range kernels. The $O(1)$ algorithm was also extended to include Gaussian $\phi(s)$ by locally approximating it using polynomials. More recently, Yang et al. \cite{Wang_bf} have proposed a $O(1)$ algorithm for arbitrary range and spatial kernels by extending the bilateral filtering method of Durand et al. \cite{Durand}. Their algorithm is based on a piecewise-linear approximation of the bilateral filter obtained by quantizing $\phi(s)$.   

In this paper, we extend the $O(1)$ algorithm of Porikli to provide an exact implementation of the bilateral filter, using trigonometric range kernels. Our main observation that trigonometric functions share a common property of polynomials which allows one to ``linearize'' the otherwise non-linear bilateral filter. The common property is that the translate of a polynomial (resp. trigonometric function) is again a polynomial (resp. trigonometric function), and importantly, of the same degree. By fixing $\phi(s)$ to be a trigonometric function, we show how this self-shiftable property can be used to (locally) linearize the bilateral filter. This  is the crux of the idea that was used for deriving the $O(1)$ algorithm for polynomial $\phi(s)$ in \cite{bilateralFilter_fast}.

\section{Constant-time bilateral filter}

\subsection{The main idea}

It is the presence of the term $\phi(f(\x-\y)-f(\x))$ in \eqref{BF} that makes the filter non-linear. In the absence of this term, that is, when $\phi(s)$ is constant, the filter is simply given by the averaging 
\begin{equation}
\label{averaging}
\overline{f}(\x)=\int_{\Omega} w(\y) f(\x-\y) \ d\y,
\end{equation}
where we assume $w(\x)$ to have a total mass of unity. It is well-known that \eqref{averaging} can be implemented in constant-time, irrespective of the size and shape of the filter, using the convolution-multiplication property of the (fast) Fourier transform. The number of computations required per pixel, however, depends on the size of the image in this case \cite{image_processing_book}. On the other hand, it is known that \eqref{averaging} can be realized at the cost of a constant number of operations per pixel (independent of the size of the image and the filter) using recursive algorithms. These $O(1)$ recursive algorithms are based on specialized kernels, such as the box and the hat function \cite{Heckbert,Crow,Young}, and the more general class of Gaussian-like box splines \cite{kunal_tip}. 

Our present idea is to leverage these fast averaging algorithms by expressing \eqref{BF} in terms of \eqref{averaging}, where the averaging is performed on the image and its simple pointwise transforms. Our observation is that we can do so if the range kernel is of the form
\begin{equation}
\label{cosine}
\phi(s)= \cos(\gamma s) \qquad (-T \leq s \leq T).
\end{equation}
By plugging \eqref{cosine} into \eqref{BF}, we can write the integral as
\begin{equation*}
  \cos(\gamma f(\x)) \int_{\Omega} w(\y)   \cos(\gamma f(\x-\y))  f(\x-\y) \ d\y+  \sin(\gamma f(\x))\int_{\Omega} w(\y)   \sin(\gamma f(\x-\y))  f(\x-\y) \ d\y.
\end{equation*}
This is clearly seen to be the linear combination of two spatial averages, performed on the images $\cos( \gamma f(\x)) f(\x)$
and $\sin(\gamma f(\x)) f(\x)$. Similarly, we can write the integral in \eqref{normalization} as
\begin{equation*}
  \cos(\gamma f(\x)) \int_{\Omega} w(\y)   \cos(\gamma f(\x-\y)) \ d\y+  \sin(\gamma f(\x))\int_{\Omega} w(\y)   \sin(\gamma f(\x-\y))  \ d\y.
\end{equation*}
In this case, the averaging is on the images $\cos(\gamma f(\x))$ and $\sin(\gamma f(\x))$. This is the trick that allows us to express \eqref{BF} in terms of  linear convolution filters applied to pointwise transforms of the image. 

\begin{figure} 
\centering 
\includegraphics[width=0.8\linewidth]{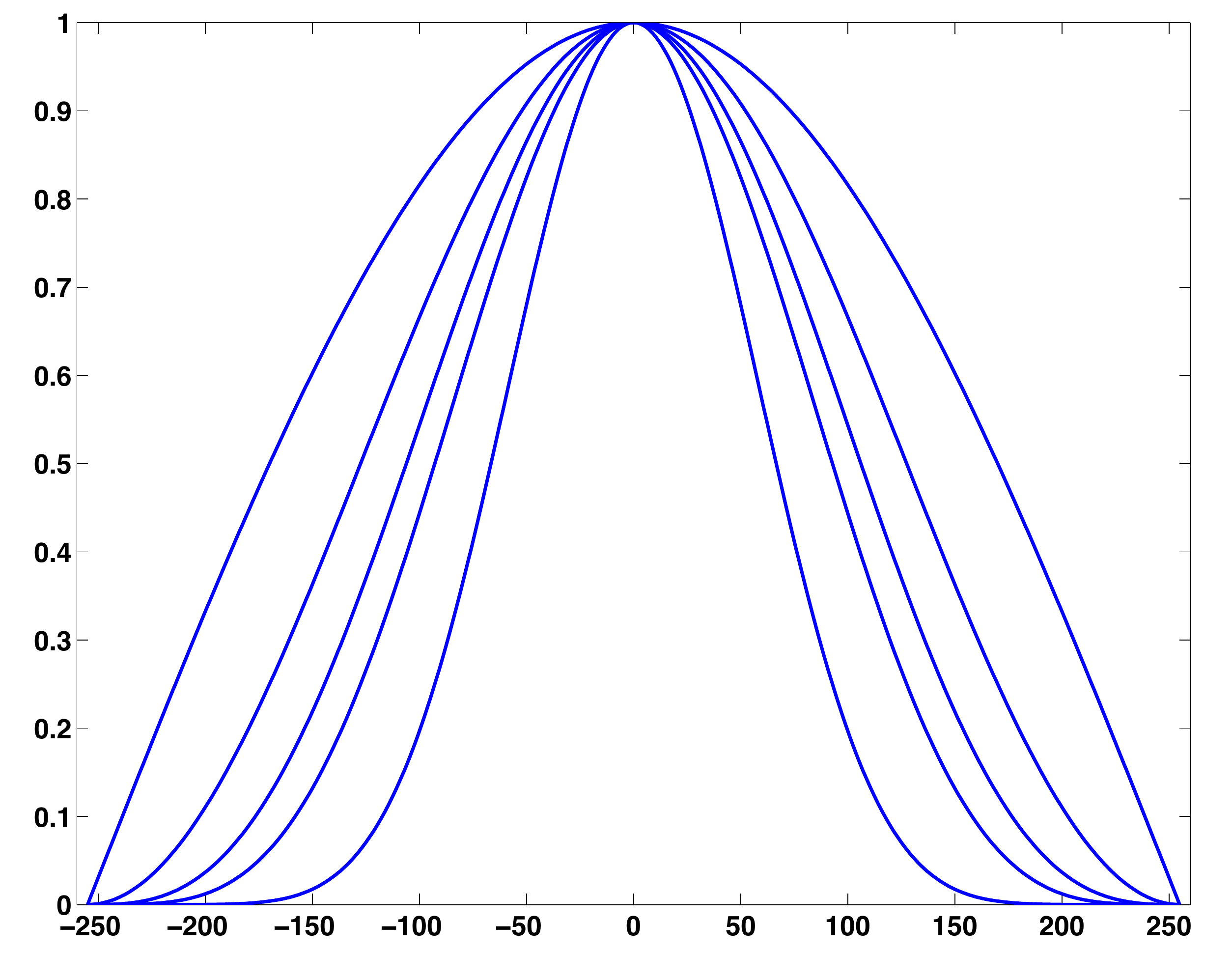}
\caption{The family of raised cosines $g(s)=[\cos (\gamma s)]^N$ over the dynamic range $-T \leq s \leq T$ as $N$ goes from $1$ to $5$ (outer to inner curves). We set $T=255$ corresponding to the maximum dynamic range of a grayscale image, and $\gamma=\pi/2T$.  They satisfy the two essential properties required to qualify as a valid range kernel of the bilateral filter---non-negativity and monotonicity (decay). Moreover, they have the remarkable property that they converge to a Gaussian (after appropriate normalization) as $N$ gets large; see \eqref{convergence}.} 
\label{raised_cosine} 
\end{figure}

Note that the domain of $\phi(s)$ is $[-T,T]$ in \eqref{cosine}. We assume here (without loss of generality) that the dynamic range of the image is within $[0,T]$. The maximum of $|f(\x)-f(\y)|$ over all $\x$ and $\y$ such that $\x-\y \in \Omega$ is within $T$ in this case. Therefore, by letting $\gamma=\pi/2T$, we can guarantee the argument $\gamma s$ of the cosine function to be within the range $[-\pi/2,\pi/2]$. The crucial point here is that the cosine function is oscillating and can assume negative values over $(-\infty,\infty)$. However, its restriction over the half-period $[-\pi/2,\pi/2]$ has two essential properties of a range kernel---it is non-negative and has a bump shape (cf. the outermost curve in Figure \ref{raised_cosine}). Note that, in practice, the bound on the local variations of intensity could be much lower than $T$. 

\subsection{General trigonometric kernels}

The above idea can easily be extended to more general trigonometric functions of the form $\phi(s)= a_0+a_1\cos(\gamma s)+ \cdots + a_N\cos(N\gamma s)$. This is most conveniently done by writing $\phi(s)$ in terms of complex exponentials, namely as
\begin{equation}
\label{kernel}
\phi(s)=\sum_{|n| \leq N} c_n \exp\big (j n \gamma s \big).
\end{equation}
The coefficients $c_n$ must be real and symmetric, since $\phi(s)$ is real and symmetric. Now, using the addition-multiplication property of exponentials, we can write
\begin{equation*}
\phi(f(\x-\y)-f(\x))=\sum_{|n| \leq N} d_n(x) \ \exp \big(j n \gamma f(\x-\y)\big)
\end{equation*}
where $d_n(\x)=c_n \exp \big(-j n \gamma f(\x)\big)$. Plugging this into \eqref{BF}, we immediately see that
\begin{equation}
\label{final-form}
\tilde{f}(\x) =  \frac{ \sum_{|n| \leq N}  \ d_n(\x) \ \overline{g_n}(\x)}{\sum_{|n| \leq N} \ d_n(\x) \  \overline{h_n}(\x)}
\end{equation}
where $h_n(\x)=\exp \big(j n \gamma  f(\x) \big)$, and  $g_n(\x)= f(\x)h_n(\x)$. We refer to $h_n(\x)$ and $g_n(x)$ as the \textit{auxiliary images}, and $N$ as the \textit{degree} of the kernel. 

The above analysis gives us the following $O(1)$ algorithm for the bilateral filter: We first set up the auxiliary images and the coefficients $d_n(\x)$ from the input image. We then average each of the auxiliary images using a $O(1)$ algorithm (this can be done in parallel). The samples of the filtered image is then given by the simple sum and division in \eqref{final-form}. In particular, for an image of size $M \times M$, we can compute the spatial averages for any arbitrary $w(\x)$ at the cost of $O(M^2 \log_2 M)$ operations using the Fourier transform. As mentioned earlier, this can  further be reduced to a total of $O(M^2)$ operations using specialized spatial kernels \cite{Heckbert,Crow,kunal_tip}. 

\subsection{Raised cosines}

\begin{figure} 
\centering 
\includegraphics[width=0.8\linewidth]{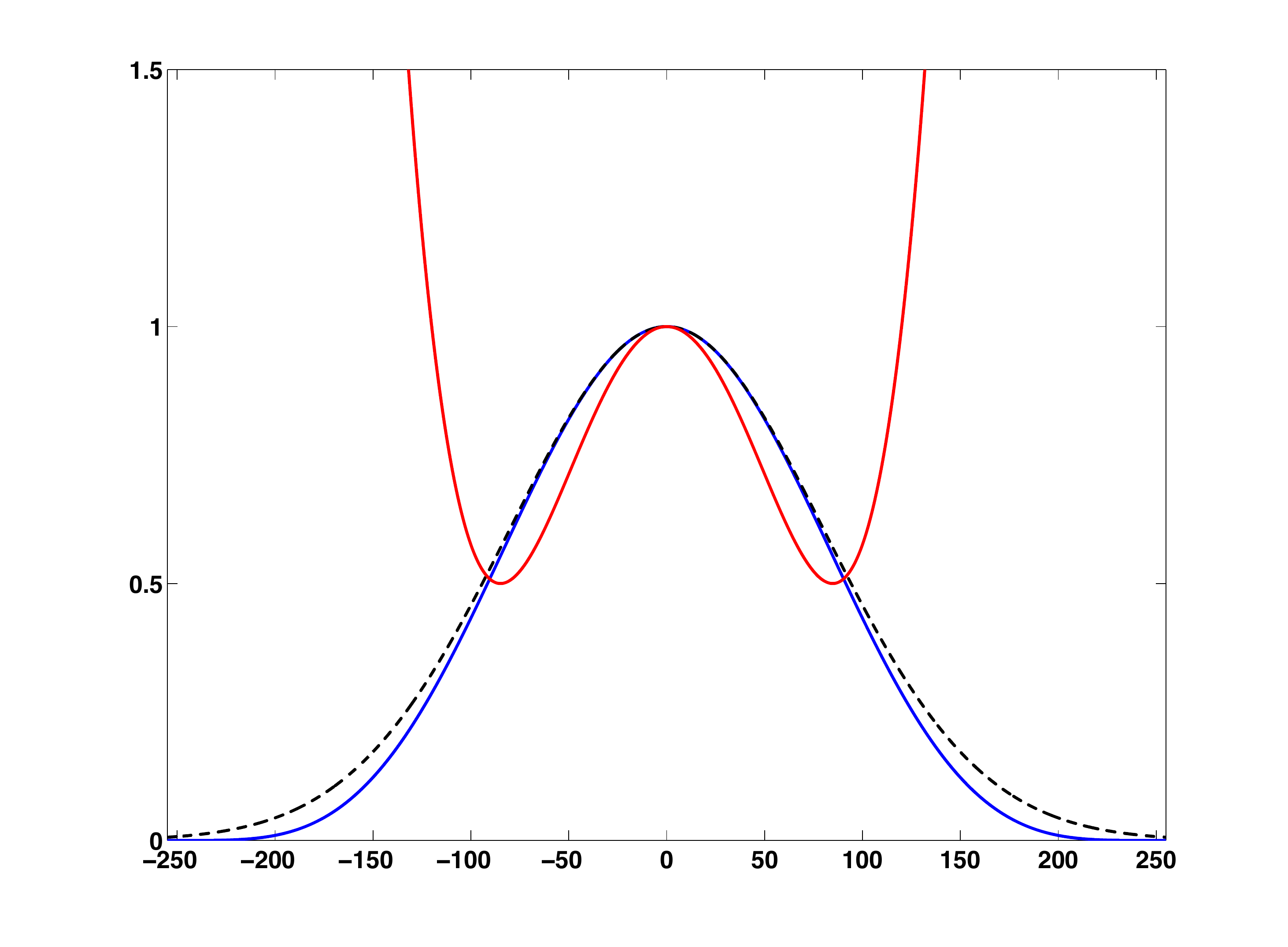}
\caption{Approximation of the Gaussian $\exp(-x^2/2\sigma^2)$ (dashed black curve) over the interval $[-255,255]$ using the Taylor polynomial (solid red curve) and the raised cosine (solid blue curve). We set $\sigma=80$, and use $N=4$ for the raised cosine in \eqref{convergence}. The raised cosine is of the form $a_0+a_1\cos(2\theta)+a_2 \cos(4 \theta)$ in this case. We use a $3$-term Taylor polynomial of the form $b_0+b_1x^2+b_2x^4$. It is clear that the raised cosine offers a much better approximation than its polynomial counterpart. In particular, note how the polynomial blows up beyond $|x| > 100$.} 
\label{approximations} 
\end{figure} 

We now address the fact that $\phi(s)$ must have some additional properties to qualify as a valid range kernel (besides being symmetric). Namely, $\phi(s)$ must be non-negative, and must be monotonic in that $\phi(s_1) \leq \phi(s_2)$ whenever $|s_1| > |s_2|$. In particular, it must have a peak at the origin. This ensures that large differences in intensity gets more penalized than small differences, and that \eqref{BF} behaves purely as a spatial filter in a region having uniform intensity. Moreover, one must also have some control on the variance (effective width) of $\phi(s)$. We now address these design problems in order. 

The properties of symmetry, non-negativity, and monotonicity are simultaneously enjoyed by the family of \textit{raised cosines} of the form
\begin{equation*}
\phi(s)=\big [ \cos(\gamma s) \big]^N \qquad (-T \leq s \leq T).
\end{equation*}
Writing $\cos \theta=(e^{j\theta}+e^{-j\theta})/2$, and applying the binomial theorem, we see that
\begin{equation*}
\phi(s)=\sum_{n=0}^N \ 2^{-N} \binom{N}{n} \exp \big(j (2n-N) \gamma s \big).
\end{equation*}
This expresses the raised cosines as in \eqref{kernel}, though we have used a slightly different summation. Since  $\phi(s)$ has a total of $(N+1)$ terms, this gives 
a total of $2(N+1)$ auxiliary images in \eqref{final-form}. The central term $n=N/2$ is constant when $N$ is even, and we have one less auxiliary image to process in this case.

\subsection{Approximation of Gaussian kernels}

Figure \ref{raised_cosine} shows the raised cosines of degree $N=1$ to $N=5$. It is seen that $\phi(s)$ become more Gaussian-like over the half-period $[-\pi,\pi]$ with the increase in $N$. The fact, however, is that $\phi(s)$ converges pointwise to zero at all points as N gets large, excepting for the node points $0, \pm \pi, \pm 2 \pi, \ldots$. This problem can nevertheless be addressed by suitably scaling the raised coinse. The precise result is given by the following  pointwise convergence:
\begin{equation}
\label{convergence}
\lim_{N \longrightarrow \infty} \left[\cos \left( \frac{\gamma s}{\sqrt N}\right)\right]^N=\exp\left(-\frac{\gamma^2 s^2}{2}\right).
\end{equation}

\begin{proof}
Note that Taylor's theorem with remainder tells us that if $f(x)$ is sufficiently smooth, then $f(x)=\sum_{k=0}^{m-1} x^k f^{(k)}(0)/k! + x^m f^{(m)}(\theta)/m!$, where $\theta$ is some 
number between $0$ and $x$. Applied to the cosine function, we have $\cos(x)=1-x^2/2+x^4 \cos \theta/24$. In other words, $\cos(x)=1-x^2/2+r(x)$, where $|r(x)| \lesssim x^4$ (we write $f(x) \lesssim g(x)$ to signify that $f(x) \leq C g(x)$ for some absolute constant $C$, where $C$ is independent of $x$). Using this estimate, along with the 
binomial theorem, we can write
\begin{align*}
 \left[\cos \left( \frac{\gamma s}{\sqrt N}\right)\right]^N =\left(1- \frac{\gamma^2 s^2}{2 N}\right)^N+\sum_{k=1}^N \binom{N}{k} r(s,N)^k \left(1- \frac{\gamma^2 s^2}{2 N}\right)^{N-k},
\end{align*}
where $|r(s,N)| \lesssim  s^4/N^2$. We are almost done since it is well-known that $(1+x/N)^N$ approaches $\exp(x)$ as $N$ gets large. To establish \eqref{convergence}, all we need to 
show is that, for any fixed $s$, the residual terms can be made negiligibly small simply by setting $N$ large.

Now note that if $|s| \lesssim N^{1/2}$, then the magnitude of $(1-\gamma^2 s^2/2 N)$ is within unity, and, on the other hand, $s^4/N< 1$ when $|s|< N^{1/4}$. Thus, given any fixed $s$, we set $N$ to be large enough 
so that $s$ satisfies the above bounds. Then, following the trivial inequality $\binom{N}{k} < N^k$, we see that the modulus of the residual is
\begin{align*}
\lesssim \sum_{k=1}^N N^k \left(\frac{s^4}{N^2}\right)^k \lesssim  N \left(\frac{s^4}{N}\right)^N \lesssim \frac{1}{N},
\end{align*}
provided that $|s| < L_N=(N^{1-2/N})^{1/4}$. This can clearly be achieved by increasing $N$, since $L_N$ is monotonic in $N$.
\end{proof}

We have seen that raised cosines of sufficiently large order provide arbitrarily close approximations of the Gaussian. The crucial feature about \eqref{convergence} is that the rate of convergence is much faster than that of Taylor polynomials, which were used to approximate the Gaussian range kernel in \cite{bilateralFilter_fast}. In particular, we can obtain an approximation comparable to that achieved using polynomials using fewer number of terms. This is important from the practical standpoint. In Figure \ref{approximations}, we consider the target Gaussian kernel $\exp(-s^2/2\sigma^2)$, where $\sigma=80$. We approximate this using the raised cosine of degree 4, which has 3 terms. We also plot the polynomial corresponding to the $3$-term Taylor expansion of the Gaussian, which is used in for approximating the Gaussian in \cite{bilateralFilter_fast}. It is clear that the approximation quality of the raised cosine is superior to that offered by a Taylor polynomial having equal number of terms. In particular, note that the Taylor approximation does not automatically offer the crucial monotonic property.

\begin{table}[!htbp]
\caption{$N_0$ is the minimum degree of the raised cosine required to approximate a Gaussian of standard deviation $\sigma$ on the interval $[-255,255]$. The estimate $\lceil (\gamma \sigma)^{-2} \rceil$ is also shown.} 
\label{spread} 
\centering
\begin{tabular}{|c|c|c|c|c|c|c|c|}
\hline
$\sigma$   & 200 &  150 &  100  & 80 &  60  &  50 & 40  \\
\hline
$N_0$  & 1  &  2  &  3 &  4 &  5  &  7 & 9   \\
\hline
$\lceil (\gamma \sigma)^{-2} \rceil$  & 1  &  2  &  3 &  5 &  8  &  11 & 17   \\
\hline
\end{tabular}
\end{table}

\subsection{Control of the width of range kernel}

The approximation in \eqref{convergence} also suggests a means of controlling the variance of the raised cosine, namely, by controlling the variance of the target Gaussian. The target Gaussian (with normalization) has a fixed variance of $\gamma^{-2}$. This can be increased simply by rescaling the argument of the cosine in \eqref{convergence} by some $\rho > 1$. In particular, for sufficiently large $N$, 
\begin{equation}
\label{convergence1}
 \left[\cos \left( \frac{\gamma s}{\rho \sqrt N}\right)\right]^N \approx  \exp\left(-\frac{s^2}{2 \rho^2 \gamma^{-2}}\right).
\end{equation}
The variance of the target Gaussian (again with normalization) has now increased to $\rho^2 \gamma^{-2}$. A fairly accurate estimate of the variance of the raised cosine is therefore $\sigma^2 \approx \rho^2 \gamma^{-2}$. In particular, we can increase the variance simply by setting $\rho=\gamma \sigma$ for all $\sigma > \gamma^{-2}$, provided $N$ is large enough. 

Bringing down the variance below $\gamma^{-2}$, on the other hand, is more subtle. This cannot be achieved simply by rescaling with $\rho <1$ on account of the oscillatory nature of the cosine. For instance, setting $\rho < 1$ can cause $\phi(s)$ to become non-negative, or loose its monotonicity. The only way of doing so is by increasing the degree of the cosine (cf. Figure \ref{raised_cosine}). In particular, $N$ must be large enough so that the argument of  $\cos(\cdot)$ is within $[-\pi/2,\pi/2]$ for all $-T \leq s \leq T$. This is the case if
\begin{equation*}
N \geq \rho^{-2} \approx  (\gamma \sigma)^{-2}.
\end{equation*}
In other words, to approximate a Gaussian having a small variance $\sigma$, $N$ must at least be as large as $N_0 \approx (\gamma \sigma)^{-2}$. The bound is quite tight for large $\sigma$, but is loose when $\sigma$ is small. We empirically determined $N_0$ for certain values of $\sigma$ for the case $T=255$, some of which are given in Table \ref{spread}. It turned out to be much lower than the estimate $(\gamma \sigma)^{-2}$ when $\sigma$ is small. For a fixed setting of $T$ (e.g., for grayscale images), this suggests the use of a lookup table for determining $N_0$ for small $\sigma$ on-the-fly.

The above analysis leads us to an $O(1)$ algorithm for approximating the Gaussian bilateral filtering, where both the spatial and range filters are Gaussians. The steps are summarized in Algorithm \ref{algo_bilateral}.

 \begin{algorithm}[!htb]
 \caption{Fast $O(1)$ bilateral filtering for the Gaussian kernel}
 \label{algo_bilateral}
 \begin{algorithmic}
      \State \textbf{Input}: Image $f(\x)$, dynamic range $[-T,T]$, $\sigma_s^2$ and $\sigma^2_r$ for the spatial and range filters. 
      \State 1. Set $\gamma=\pi/2T$, and $\rho=\gamma \sigma_r$.
      \State 2. If $\sigma_r > \gamma^{-2}$, pick any large $N$. Else, set $N=(\gamma \sigma_r)^{-2}$, or use a look-up table to fix $N$.
      \State 3. Set $h_n(\x)=\exp \left(j \gamma (2n-N) f(\x)/\rho \sqrt N \right)$ and $g_n(\x)= f(\x) h_n(\x)$, and the coefficients $d_n(\x)=2^{-N} \binom{N}{n} \exp \left(- j \gamma (2n-N) f(\x)/\rho \sqrt N \right)$.
      \State 4. Use an $O(1)$ algorithm to filter $h_n(\x)$ and $g_n(\x)$ with a Gaussian of variance $\sigma_s^2$ to get $\overline{h_n}(\x)$ and $\overline{g_n}(\x)$. 
      \State 5. Set $\tilde{f}(\x)$ as the ratio of $\sum_{n=0}^N  \ d_n(\x) \overline{g_n}(\x)$ and $\sum_{n=0}^N d_n(\x)  \overline{h_n}(\x)$.
      \State \textbf{Return}: Filtered image $\tilde{f}(\x)$.
 \end{algorithmic}
 \end{algorithm}

\section{Experiments}

We implemented the proposed algorithm for Gaussian bilateral filtering in Java on a Mac OS X  $2\times$ Quad core 2.66 GHz machine, as an ImageJ plugin. We used multi-threading for computing the spatial averages of the auxiliary images in parallel. A recursive $O(1)$ algorithm was used for implementing the Gaussian filter in space domain \cite{image_processing_book}. The average times required for processing a $720 \times 540$ grayscale image using our algorithm are shown in Table \ref{timing}. We repeated the experiment for different variances of the Gaussian range kernel, and at different spatial variances. As seen from the table, the processing time is quite fast compared to a direct implementation of the bilateral  filter, which requires considerable time depending on the size of the spatial filter. For instance,  a direct implementation of the filter on a $512 \times 512$ image required $4$ seconds for $\sigma_s$ as low as $3$ on our machine (using discretized Gaussians supported on $[-3 \sigma,3 \sigma]^2$), and this climbed up to almost $10$ seconds for $\sigma_s=10$. As is seen from Table \ref{timing}, the processing time of our algorithm, however, suddenly shoots up for narrow Gaussians with $\sigma_r <15$. This is due to the large $N$ required to approximate the Gaussian in this regime (cf. Table \ref{spread}). We have figured out an approximation scheme for further accelerating the processing for very small $\sigma_r$, without appreciably degarding the final output. Discussion of this method is however beyond the present scope of the paper.

\begin{figure} 
\centering 
\includegraphics[width=1.0\linewidth]{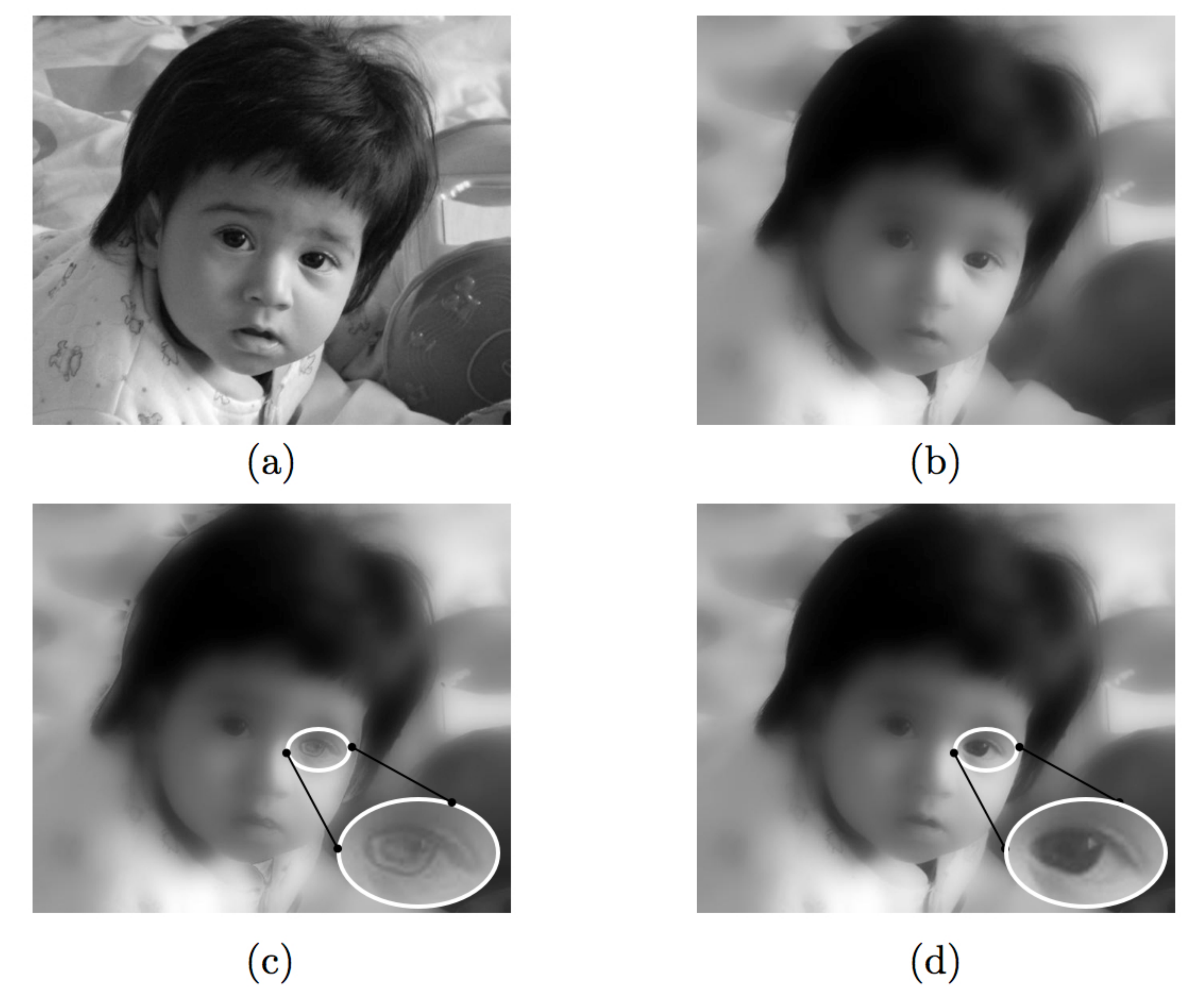}
\caption{Comparison of various implementations of the Gaussian bilateral filter on the grayscale image \textit{Isha} of size $600 \times 512$. The filter settings are $\sigma_s=15$ and $\sigma_r=80$. (a) Original image; (b) Direct implementation of the bilateral filter; (c) Output obtained using polynomial kernel \cite{bilateralFilter_fast}; and (d) Output of our algorithm. Note the strange artifacts in (c), particularly around the right eye (see zoomed insets). This is on account of the distortion caused by the polynomial approximation shown in Figure \ref{approximations}. The standard deviation of the error between (b) and (c) is $6.5$, while that between (b) and (d) is $1.2$.} 
\label{results} 
\end{figure}

We next tried a visual comparisonof the ouput of our algorithm with the algorithm in \cite{bilateralFilter_fast}. In Figure \ref{results}, we compare the outputs of the two algorithms with the direct implementation, on a natural grayscale image. As is clearly seen from the processed images, our result resembles the exact output very closely. The result obtained using the polynomial kernel, on the other hand, shows strange artifacts. The difference is also clear from the standard deviation of the error between the exact output and the approximations. We note, however, that the execution time of the polynomial method is slightly lower than that of our method, since it requires half the number of auxiliary images for a given degree. 

\begin{figure}
\centering
\begin{tabular}{cc}
\includegraphics[width=0.5\linewidth]{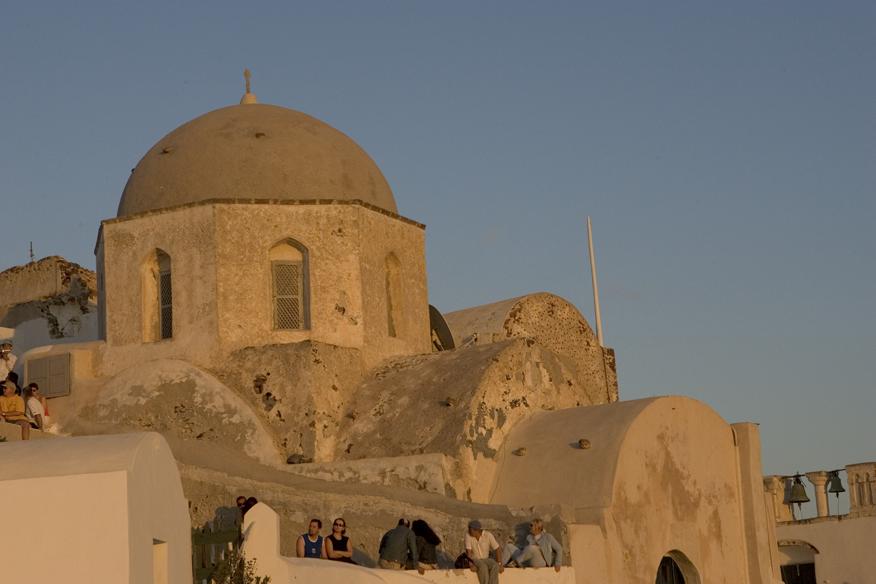} &
\includegraphics[width=0.5\linewidth]{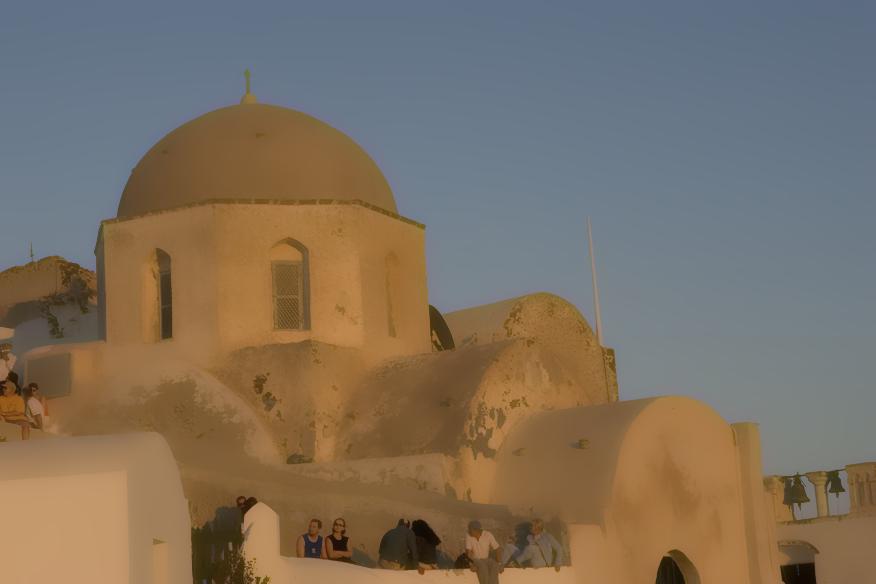} \\
\includegraphics[width=0.5\linewidth]{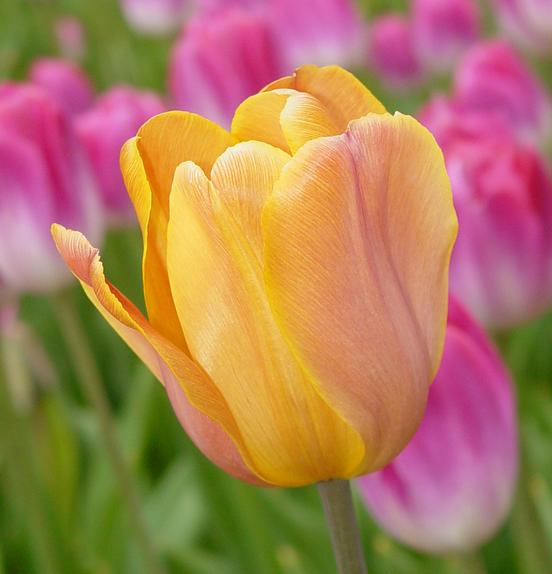} &
\includegraphics[width=0.5\linewidth]{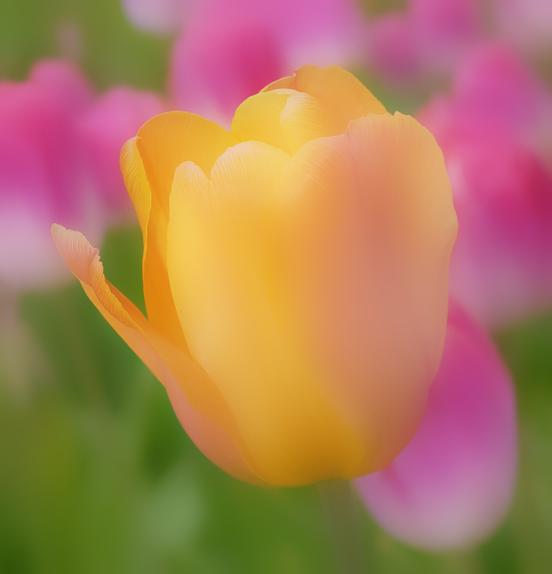}
\end{tabular}
\caption{Results on the color images \textit{Greekdome} and \textit{Tulip}, using our implementation of the Gaussian bilateral filter. The original image is on the left, and the processed image is on the right. In either case, the red, green, and blue channels were processed independently. We used $\sigma_s=10$ and $\sigma_r=20$ for \textit{Greekdome}, and  $\sigma_s=20$ and $\sigma_r=60$ for \textit{Tulip}. (Images courtsey of Sylvain Paris and Fr\'edo Durand).}
\label{results_color}
\end{figure}

We also tested our implementation of the Gaussian bilateral filter on color (RGB) images. We tried a naive processing, where each of the three color channels were processed independently. The results on a couple of images are shown in Figure \ref{results_color}. The Java source code can be downloaded from the web at http://bigwww.epfl.ch/algorithms/bilateral-filter.

\begin{table}[!htbp]
\caption{The time in milliseconds required for processing a grayscale image of size $720 \times 540$ pixels using our algorithm. The processing was done on a Mac OS X, $2\times$ quad core 2.66 GHz machine, using multithreading.} 
\label{timing} 
\centering
\begin{tabular}{|c|c|c|c|c|c|c|c|c|c|c|}
\hline
$\sigma_r \rightarrow$   & 10 & 20 & 30 & 40 & 50 & 60 & 70 & 80  & 90 & 100 \\
\hline
$\sigma_s=10$ & 3604  & 452 &  195 & 120 & 74 & 61 & 49 & 34 & 32 & 27 \\
\hline
$\sigma_s=100$  & 3755 & 482 & 217 & 127 & 89 & 69 & 54 & 43 & 37 & 28   \\
\hline
\end{tabular}
\end{table}

\section{Discussion}

We presented a general method of computing the bilateral filter in constant-time using trigonometric range kernels. Within this framework, we showed how feasible range kernels can be realized using the family of raised cosines. The highlights of our approach are the following: 
\vspace{1.5mm}

\noindent $\bullet$ \textbf{Accuracy}. Our method is exact, at least for the family of raised cosines. It does not require the quantization of the range kernel, as is the case in \cite{Durand,Wang_bf}. Moreover, note that the auxiliary images in \eqref{final-form} have the same dynamic range as the input image irrespective of the degree $N$. This is unlike the situation in \cite{bilateralFilter_fast}, where the dynamic range of the auxiliary images grow exponentially with the $N$. This makes the computations susceptible to numerical errors for large $N$.

\noindent $\bullet$ \textbf{Speed}. Besides having $O(1)$ complexity, our algorithm can also be implemented in parallel. This allows us to further accelerate its speed. 

\noindent $\bullet$  \textbf{Approximation property}. Trigonometric functions yield better (local) approximation of Gaussians than polynomials. In particular, we showed that by using a particular class of raised cosines, we can obtain much better approximations of the Gaussian range kernel than that offered by the Taylor polynomials in \cite{bilateralFilter_fast}. The final output is artifact-free and resembles the true output very closely. The only flip side of our approach (this is also the case with \cite{bilateralFilter_fast}, as noted in \cite{Wang_bf}) is that a large number of terms are required to approximate very narrow Gaussians over large intervals. 

\noindent $\bullet$ \textbf{Space-variant extension}. The spatial kernel in \eqref{BF} can be changed from point-to-point within the image to control the amount of smoothing (particularly in homogenous regions), while the range kernel is kept fixed. Thanks to \eqref{final-form}, this can be done simply by  computing the space-variant averages of each auxiliary image. The good news is that this can also be realized for a $M \times M$ image at the cost of $O(M^2)$ operations, using particular spatial kernels. This includes the two-dimensional box and hat filter \cite{Heckbert,Crow}, and the more general class of Gaussian-like box splines in \cite{kunal_tip}.

\section{Acknowledgement}

The authors thank Ayush Bhandari for his help with the insets in Figure \ref{results}, and also Sagnik Sanyal for providing the image used in the same figure.  

\bibliographystyle{plain}
\bibliography{bilateral}

\begin{thebibliography}{10}

\bibitem{Bennet}
E.P. Bennett, J.L. Mason, and L.~McMillan.
\newblock Multispectral bilateral video fusion.
\newblock {\em {IEEE} Transactions on Image Processing}, 16:1185--1194, 2007.

\bibitem{Baudes}
A.~Buades, B.~Coll, and J.M. Morel.
\newblock A review of image denoising algorithms, with a new one.
\newblock {\em Multiscale {M}odeling and {S}imulation}, 4:490--530, 2005.

\bibitem{kunal_tip}
K.N. Chaudhury, A.~Mu{\~{n}}oz-Barrutia, and M.~Unser.
\newblock Fast space-variant elliptical filtering using box splines.
\newblock {\em {IEEE} Transactions on Image Processing}, 19:2290--2306, 2010.

\bibitem{Crow}
F.~C. Crow.
\newblock Summed-area tables for texture mapping.
\newblock {\em {ACM} Siggraph}, 18:207--212, 1984.

\bibitem{Durand}
F.~Durand and J.~Dorsey.
\newblock Fast bilateral filtering for the display of high-dynamic-range
  images.
\newblock {\em {ACM} Siggraph}, 21:257--266, 2002.

\bibitem{Heckbert}
P.S. Heckbert.
\newblock Filtering by repeated integration.
\newblock {\em International Confernece on Computer Graphics and Interactive
  Techniques}, 20(4):315--321, 1986.

\bibitem{Paris}
S.~Paris and F.~Durand.
\newblock A fast approximation of the bilateral filter using a signal
  processing approach.
\newblock {\em European Conference on Computer Vision}, pages 568--580, 2006.

\bibitem{PeronaMalik}
P.~Perona and J.~Malik.
\newblock Scale-space and edge detection using anisotropic diffusion.
\newblock {\em {IEEE} Transactions on Pattern Analysis and Machine
  Intelligence}, 12(7):629--639, 1990.

\bibitem{Pham}
T.Q. Pham and L.J. van Vliet.
\newblock Separable bilateral filtering for fast video preprocessing.
\newblock {\em {IEEE} International Conference on Multimedia and Expo}, pages
  1--4, 2005.

\bibitem{bilateralFilter_fast}
F.~Porikli.
\newblock Constant time {$O(1)$} bilateral filtering.
\newblock {\em {IEEE} Conference on Computer Vision and Pattern Recognition},
  pages 1--8, 2008.

\bibitem{Ramanath}
R.~Ramanath and W.~E. Snyder.
\newblock Adaptive demosaicking.
\newblock {\em Journal of Electronic Imaging}, 12:633--642, 2003.

\bibitem{bilateralFilter}
C.~Tomasi and R.~Manduchi.
\newblock Bilateral filtering for gray and color images.
\newblock {\em {IEEE} International Conference on Computer Vision}, pages
  839--846, 1998.

\bibitem{Weiss}
B.~Weiss.
\newblock Fast median and bilateral filtering.
\newblock {\em {ACM} Siggraph}, 25:519--526, 2006.

\bibitem{VideoAbstraction}
H.~Winnem\"{o}ller, S.~C. Olsen, and B.~Gooch.
\newblock Real-time video abstraction.
\newblock {\em {ACM} Siggraph}, pages 1221--1226, 2006.

\bibitem{Xiao}
J.~Xiao, H.~Cheng, H.~Sawhney, C.~Rao, and M.~Isnardi.
\newblock Bilateral filtering-based optical flow estimation with occlusion
  detection.
\newblock {\em European Conference on Computer Vision}, pages 211--224, 2006.

\bibitem{Wang_bf}
Q.~Yang, K.-H. Tan, and N.~Ahuja.
\newblock Real-time ${O}(1)$ bilateral filtering.
\newblock {\em {IEEE} Conference on Computer Vision and Pattern Recognition},
  pages 557--564, 2009.

\bibitem{Yang_BLF_stereo}
Q.~Yang, L.~Wang, R.~Yang, H.~Stewenius, and D.~Nister.
\newblock Stereo matching with color-weighted correlation, hierarchical belief
  propagation and occlusion handling.
\newblock {\em {IEEE} Transaction on Pattern Analysis and Machine
  Intelligence}, 31:492--504, 2009.

\bibitem{image_processing_book}
I.~Young, J.~Gerbrands, and L.~van Vliet.
\newblock {\em Fundamentals of {I}mage {P}rocessing}.
\newblock Delft PH Publications, 1995.

\bibitem{Young}
I.T. Young and L.J. van Vliet.
\newblock Recursive implementation of the {G}aussian filter.
\newblock {\em Signal Processing}, 44(2):139--151, 1995.

\end{thebibliography}

\end{document}